%% file: main.tex
\documentclass[runningheads]{llncs}

\usepackage{amsmath, amssymb, amsfonts}
\usepackage{algorithm, algorithmic}
\usepackage{balance}
\usepackage{cite}
\usepackage{enumitem}
\usepackage{textcomp}
\usepackage{xcolor}
\DeclareMathOperator*{\argmax}{arg\,max}

\usepackage{graphics}
\usepackage{graphicx}
\graphicspath{{figures/}}

\begin{document}
\title{Real-time Detection of Practical Universal Adversarial Perturbations\thanks{Kenneth T. Co is supported in part by the DataSpartan research grant DSRD201801.}}
\titlerunning{Real-time Detection of UAPs}
%
\author{Kenneth T. Co\inst{1,2}\orcidID{0000-0003-2766-7326} \and
Luis Mu\~noz-Gonz\'alez\inst{1}\orcidID{0000-0001-6093-5922} \and
Leslie Kanthan\inst{2} \and
Emil C. Lupu\inst{1}\orcidID{0000-0002-2844-3917}}
\authorrunning{K. Co et al.}
%
\institute{Imperial College London, London SW7 2AZ, United Kingdom
\email{\{k.co,l.munoz,e.c.lupu\}@imperial.ac.uk}\\
\and DataSpartan, London EC2Y 9ST, United Kingdom\\
\email{leslie@dataspartan.com}}
\maketitle              
\begin{abstract}
Universal Adversarial Perturbations (UAPs) are a prominent class of adversarial examples that exploit the systemic vulnerabilities and enable physically realizable and robust attacks against Deep Neural Networks (DNNs). UAPs generalize across many different inputs; this leads to realistic and effective attacks that can be applied at scale. In this paper we propose HyperNeuron, an efficient and scalable algorithm that allows for the real-time detection of UAPs by identifying suspicious neuron hyper-activations. Our results show the effectiveness of HyperNeuron on multiple tasks (image classification, object detection), against a wide variety of universal attacks, and in realistic scenarios, like perceptual ad-blocking and adversarial patches. HyperNeuron is able to simultaneously detect both adversarial mask and patch UAPs with comparable or better performance than existing UAP defenses whilst introducing a significantly reduced latency of only 0.86 milliseconds per image. This suggests that many realistic and practical universal attacks can be reliably mitigated in real-time, which shows promise for the robust deployment of machine learning systems.
\end{abstract}

\input{sections/1-introduction}
\input{sections/2-background}
\input{sections/3-detector}
\input{sections/4-imagenet}
\input{sections/5-practice}
\input{sections/6-comparison}
\input{sections/7-adaptive}
\input{sections/8-adblocking}
\input{sections/X-conclusion}
\clearpage

\bibliographystyle{splncs04}
\bibliography{main}
\end{document}

%% file: sections/1-introduction.tex
\section{Introduction}
Advances in computation and machine learning have enabled Deep Neural Networks (DNNs) to become the algorithm of choice for many applications such as image classification \cite{krizhevsky2012imagenet}, real-time object detection \cite{redmon2016you}, and speech recognition \cite{hinton2012deep}. However, despite their success, DNNs remain vulnerable to adversarial examples: inputs that appear similar to genuine data, but are designed to deceive the model \cite{biggio2013evasion, szegedy2014intriguing}. It is therefore important to ensure that DNNs are robust to such attacks, which undermine the performance and trust in these algorithms.

Universal Adversarial Perturbations (UAPs) are a class of adversarial perturbations where a single perturbation causes a model to misclassify on a large set of inputs \cite{moosavi2017universal}. They present a systemic risk and a number of practical and realistic adversarial attacks on machine learning models are based on UAPs. Physically realizable universal attacks in the form of adversarial patches have been demonstrated against image classification \cite{brown2017adversarial}, person recognition \cite{thys2019fooling}, facial recognition \cite{sharif2016accessorize}, and object detection \cite{eykholt2018physical, eykholt2018robust, liu2018dpatch}. Adversarial objects created via 3D-printing can fool image classifiers \cite{athalye2017synthesizing} and LiDAR-based object detection \cite{cao2019adversarial, hau2020ghostbuster, hau2021object, tu2020physically}. Universal attacks evade perceptual ad-blocking on web pages \cite{tramer2019adversarial}. An attacker can also use UAPs to perform extremely query-efficient black-box attacks \cite{co2019procedural}. All these attacks are \textit{universal}, as their perturbations are effective across inputs and robust to changes in the input. In contrast to input-specific (``per-input'') attacks, UAPs are far more efficient as a single perturbation fools a model on a large dataset. It is this property that allows universal attacks to be a significant threat in practice. Although a number of studies have addressed the detection of adversarial examples, many focus on per-input perturbations and do not consider the specificities of UAPs. Existing methods either introduce delays or significant penalties to clean performance. In contrast, we focus here on fast and scalable methods that can detect UAPs at the rate at which attacks can be performed.

In this paper, we propose HyperNeuron, a fast and scalable method to detect universal attacks against DNNs. HyperNeuron flags inputs that hyper-activate hidden layer neurons sufficiently exceed a trusted baseline. We test this novel method against a wide variety of universal attacks. Our results show that HyperNeuron can detect both adversarial patch and adversarial mask UAPs with comparable or better performance when compared to existing UAP defenses. HypernNeuron introduces minimal latency making it usable in real-time applications such as object detection and perceptual ad-blocking. It can also be a viable pre-filter to complement more computationally expensive defense mechanisms. Overall, this paper makes the following contributions:
\begin{enumerate}
    \item We propose HyperNeuron, a fast and scalable detector that is able to simultaneously detect both adversarial mask and adversarial patch universal attacks with extremely low latency.
    \item We show that UAPs exhibit a trade-off between detectability and universality, even for defense-aware attacks. Our results provide a more general methodology for comparing and evaluating existing and future universal attacks and defenses.
    \item We highlight the role of UAPs in existing practical attacks on DNNs, and show that HyperNeuron can successfully detect attacks in realistic scenarios, such as adversarial patches against image classification and universal attacks on perceptual ad-blocking.
\end{enumerate}

The rest of this paper is organized as follows. Sec.~2 introduces adversarial examples and existing universal attacks. Sec.~3 motivates and describes HyperNeuron, our detection algorithm. Sec.~4 evaluates HyperNeuron on a widely-used large scale image dataset against adversarial patch and mask attacks. Sec.~5 to 8 consider the performance of HyperNeuron in practical settings and against defense-aware attacks. Finally, Sec.~9 summarizes our findings.

%% file: sections/2-background.tex
\begin{figure}[t]
\centering
\includegraphics[width = 0.6 \columnwidth]{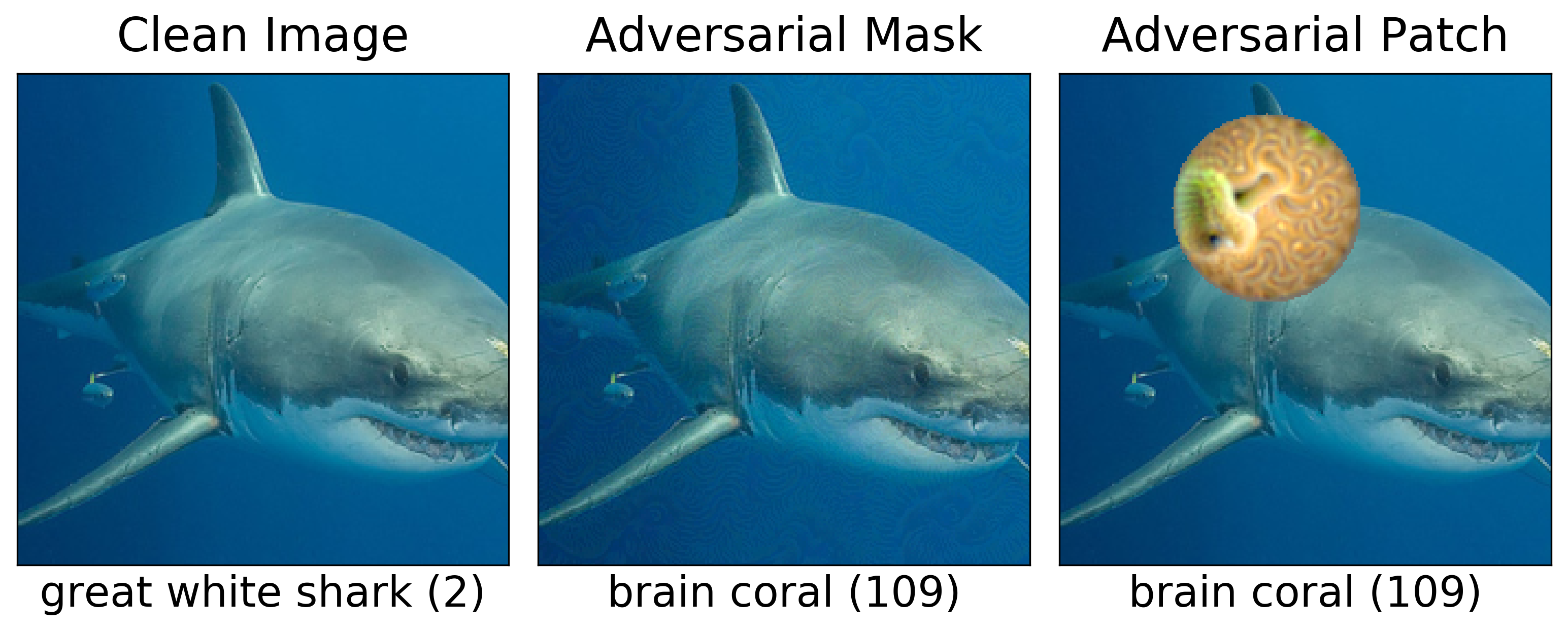}
\includegraphics[width = 0.6 \columnwidth]{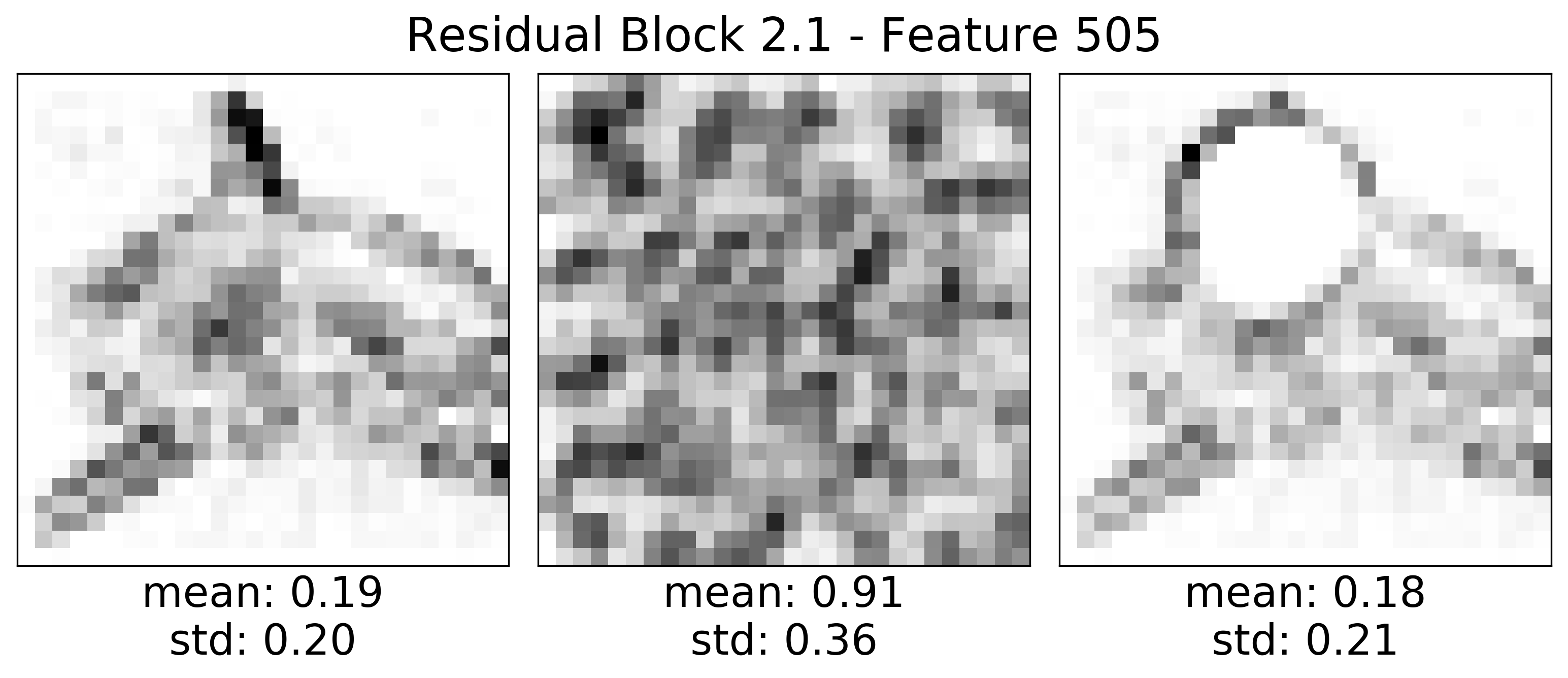}
\includegraphics[width = 0.6 \columnwidth]{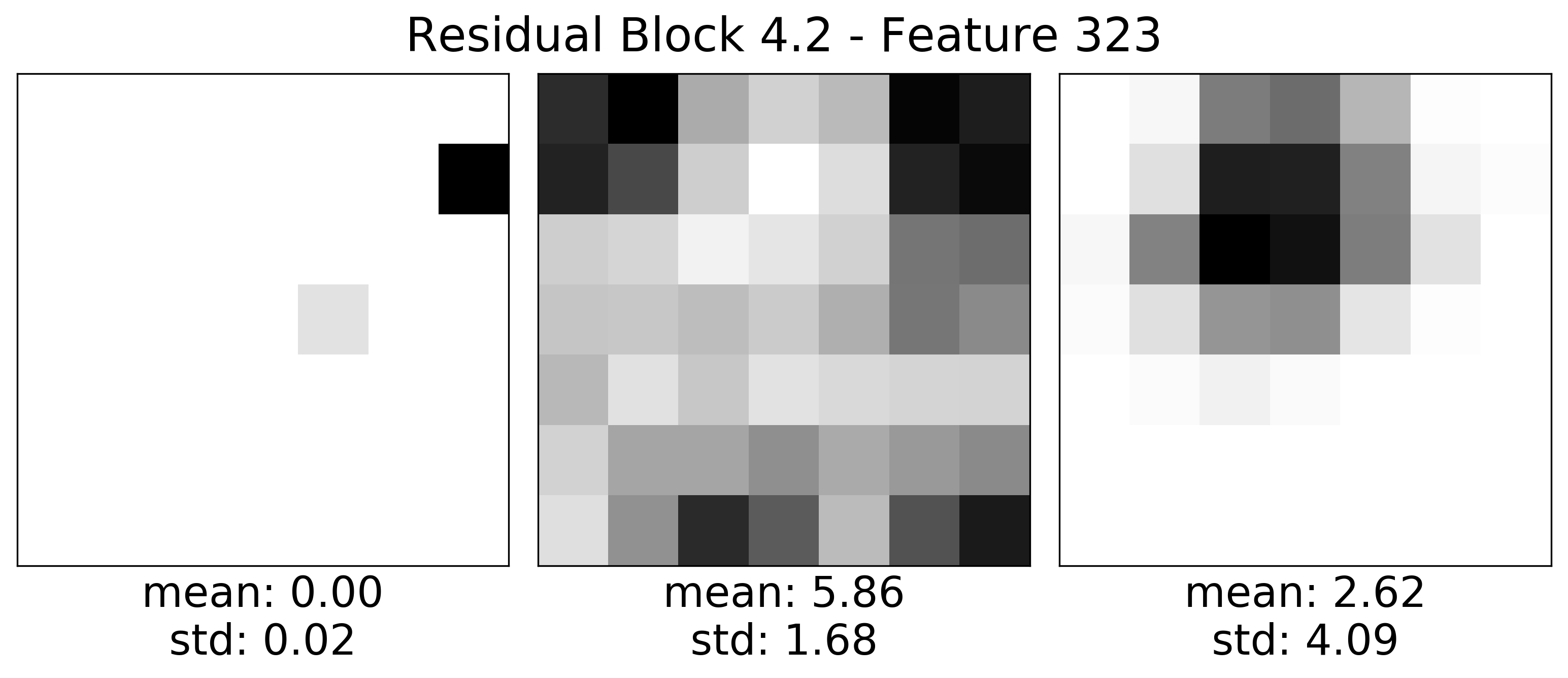}
\caption{Visualizations for an image from the ImageNet dataset (left), with an adversarial mask (center), and with an adversarial patch (right). The clean image is classified as a great white shark (class \#2) while the other two are classified as brain coral (class \#109) by a ResNet50 model. The second and third rows visualize selected features from chosen residual blocks of the model. The features' computed mean and standard deviation (std) are shown below and are noticeably larger for the perturbed inputs.}
\label{fig:intuition}
\end{figure}

\section{Background}
In classification problems, a neural network can be written as a function $F$ which takes an input $x \in \mathbb{R}^n$ and outputs a probability vector $F(x) \in \mathbb{R}^m$. The output label assigned by this model is defined by $f(x) = \argmax(F(x))$. Existing neural networks are vulnerable to adversarial examples, maliciously altered inputs designed to fool the model \cite{szegedy2014intriguing}. Let $\tau(x)$ denote the true class label of an input $x$. An adversarial example $x'$ is an input that satisfies $f(x') \neq \tau(x)$, despite $x'$ being close to $x$ according to some distance metric (implicitly, $\tau(x) = \tau(x')$). The difference $\delta = x' - x$ is referred to as an adversarial perturbation and its norm is often constrained to $\Vert \delta \Vert_p < \varepsilon$, for some $p$-norm and small $\varepsilon > 0$.

A \textbf{Universal Adversarial Perturbation} (UAP) is a perturbation $\delta \in \mathbb{R}^n$ that satisfies $f(x + \delta) \neq \tau(x)$ for sufficiently many $x \in X$ of a benign dataset $X$. UAPs show that models have systemic vulnerabilities that can be exploited regardless of the input \cite{co2019procedural, ilyas2019adversarial, co2021jacobian}. As they are optimized over multiple inputs, UAPs exploit global model sensitivities rather than per-input characteristics. UAPs are also \textit{transferable} across models \cite{moosavi2017universal}. This property is exploited in practical black-box transfer attacks \cite{papernot2016transferability, papernot2017practical, demontis2019adversarial}. In computer vision, there are two categories of UAPs: adversarial masks and adversarial patches.

\textbf{Adversarial masks} are visually-imperceptible perturbations applied to the entire image and are most often constrained with the $\ell_{\infty}$-norm. An example of this attack in practice is the attack on perceptual ad-blocking proposed by Tramer et al. \cite{tramer2019adversarial}. Adversarial masks are usually generated by directly optimizing over the model's training loss function. These direct attacks are effective, but require white-box access to the model \cite{moosavi2017universal, shafahi2018universal, co2019universal}. Effective UAP attacks can be achieved with Stochastic Gradient Descent (SGD), which uses the Projected Gradient Descent (PGD) \cite{madry2018towards} algorithm, but optimizes over batches rather than single inputs \cite{shafahi2018universal, mummadi2019defending, tramer2019adversarial}. The algorithm optimizes $\sum_i \mathcal{L}(x_i + \delta)$, where $\mathcal{L}$ is the model's training loss, $X_{\text{batch}} = \{x_i\}$ are batches of inputs, and $\delta \in \mathcal{P}$ are the valid perturbations. Updates to $\delta$ are done in mini-batches in the direction of $-\sum_i \nabla \mathcal{L}(x_i + \delta)$. The perturbation constraints take the form of an $\ell_p$-norm and appear as $\Vert \delta \Vert_{p} \leq \varepsilon$ for a chosen norm $p$ and value $\varepsilon$. SGD is a representative example of this type of attack and is the strongest version when optimizing for universality. In a \emph{targeted attack}, the attacker aims to have all inputs classified towards a specified target class and are generated by optimizing the SGD loss to favor classification in the desired target class.

UAPs can also be generated ``indirectly'' by either optimizing a proxy objective \cite{mopuri2017fast, mopuri2018generalizable, khrulkov2018art} or by exploiting a specific property of the input domain such as in procedural noise \cite{co2019procedural} or Fourier basis functions \cite{tsuzuku2019structural}. These demonstrate that UAPs can be generated under more restrictive threat assumptions, even if they are less effective than direct attacks \cite{shafahi2018universal, co2019universal}. For proxy objectives, existing work find perturbations that maximally activate hidden layer values \cite{mopuri2017fast, mopuri2018generalizable}. To improve on this formulation, the SGD-Layer attack uses the SGD optimization algorithm with this layer maximization loss \cite{co2019universal}, thus taking the best elements from both SGD and proxy objective attacks. Due to its effectiveness and its ability to be tailored to attack specific layers, we use SGD-Layer as a representative example for this class of indirect attacks.

\textbf{Adversarial patches} are 2D image patches designed to fool the model when inserted into the image. These can have different shapes and are often constrained based on the size of the patch relative to the image. Existing work has shown that these can fool image classification \cite{brown2017adversarial}, facial recognition \cite{sharif2016accessorize}, and object detection \cite{eykholt2018physical, eykholt2018robust, liu2018dpatch, thys2019fooling}. Fig.~\ref{fig:intuition} shows an example image with a mask and patch attack. Similar to adversarial masks, patches are typically optimized over the model's training loss. In contrast to masks, patches need to be robust to image transformations such as translation, rotation, and scaling. Athalye et al. propose Expectation Over Transformations (EOT) \cite{athalye2017synthesizing}, which optimizes the expectation over transformations $\mathbb{E}_{t \sim T} \mathcal{L}(t(x))$ from a distribution of valid transformations $t \sim T$. The optimization problem is solved by SGD, noting that $\nabla \mathbb{E}_{t \sim T} \mathcal{L}(t(x)) = \mathbb{E}_{t \sim T} \nabla \mathcal{L}(t(x))$. The Robust Physical Perturbation (RP$_2$) objective proposed by Eykholt et al. also optimizes an expectation over transformations \cite{eykholt2018physical, eykholt2018robust}. Their objective function includes additional terms that capture printability constraints.

%% file: sections/3-detector.tex
\section{HyperNeuron Detector}
UAPs are designed to alter the model's decision on many inputs. Thus, we expect UAPs to elicit sufficiently large activations in the hidden layers to affect the model's output across many inputs. Fig.~\ref{fig:intuition} shows an example where UAPs increase the activation values (mean) and statistical dispersion (standard deviation) in the hidden layers. We use the elevated values of these statistical metrics to design a detector for UAP attacks. We first discuss the attacker's goals and capabilities in the threat model that we consider.

\subsection{Attacker Model}
Most universal attacks require a white-box attacker who can read the model's internal weights and training data. The attacker can query the target model with any input, and aims to generate UAPs that maximize universality. We measure this with the \emph{Universal Evasion Rate} (UER) of a perturbation over the dataset:
\begin{equation}
\text{UER}(\delta) = \vert \{x \in X : \argmax F(x + \delta) \neq \tau(x) \} \vert \cdot \frac{1}{\vert X \vert}
\end{equation}\label{eq:uer}
with $\Vert \delta \Vert_{\infty} \leq \varepsilon$. We have used the $\ell_{\infty}$-norm perturbation constraint for $\delta$ as it is frequently used in the computer vision and UAP literature. It ensures that the perturbation is small and does not greatly alter the visual appearance of the image. For targeted UAPs, the attacker aims to maximize the \emph{Targeted Success Rate} (TSR):
\begin{equation}
\text{TSR}(\delta, y_{\text{tgt}}) = \vert \{x \in X : \argmax F(x + \delta) = y_{\text{tgt}} \} \vert \cdot \frac{1}{\vert X \vert}
\end{equation}\label{eq:tgt}
where $y_{\text{tgt}}$ is the target class label. We first evaluate against attacks under the assumption that no defense has been deployed. Later on in the paper we also consider an attack that deploys an adaptive strategy against our defense.

\begin{algorithm}[t]
   \caption{HyperNeuron Algorithm}
   \label{alg:detect}
\begin{algorithmic}
\REQUIRE $X_{\text{clean}}$, $X_{\text{test}}$, aggregation function $A$, layer $k$, top percentile $t$, FP $r$,
    \STATE {$\mu_k \leftarrow \text{average}_{x \in X_{\text{clean}}}{A_k(x)}$}
    \STATE {$\sigma_k \leftarrow \text{standard-deviation}_{x \in X_{\text{clean}}}{A_k(x)}$}
    \FOR {$x_i \in X_{\text{test}}$}
        \STATE {$Z_k(x_i) \leftarrow \text{z-score}({A_k(x_i)}, \mu = \mu_k, \sigma = \sigma_k)$}
        \STATE $z_{\text{top-pct}, i} = \{\text{top-percentile}(Z_k(x_i), t)\}$, top-$t$ percentile
        \STATE $z_{\text{top-avg}, i} \leftarrow \text{average}(z_{\text{top-pct}, i})$
        \IF {$z_{\text{top-avg}, i} > \theta_{k, r}$}
            \STATE $\text{flag}(x_i) = 1$
        \ELSE
            \STATE $\text{flag}(x_i) = 0$
        \ENDIF
    \ENDFOR
    \STATE \textbf{return} $\{z_{\text{top-avg}, i}\}, \{\text{flag}(x_i)\}$
\end{algorithmic}
\end{algorithm}

\subsection{HyperNeuron Detection Algorithm}
The defender begins with a deep neural network $F$ that outputs $L_k(x)$ at layer $k$ for input $x$. We assume the defender has access to a trusted clean dataset $X_{\text{clean}}$, and has chosen a layer index $k$ to observe. Selection of $k$ is described in Sec.~4.1.

\textbf{Feature Aggregation.} We consider DNNs where the output of hidden layer $L_k(x)$ has dimensions $a_k \times a_k \times d_k$. This can be thought of as a $d_k$-dimensional vector of $a_k \times a_k$ features. Fig.~\ref{fig:intuition} shows examples of these $a_k \times a_k$ features for outputs of residual blocks 2.1 and 4.2 in the ResNet50 \cite{he2016deep} model. We aggregate the statistics of each $a_k \times a_k$ feature in $L_k(x)$ to obtain an aggregated feature vector $A_k(x) \in \mathbb{R}^{d_k}$. For each input, we use the aggregation function $A_k(\cdot)$, either the mean $A_k = M_k$ or the standard deviation $A_k = S_k$. Although other statistical measures could be chosen, we show that these work well in our experiments.

The aggregation of features is needed to reduce the memory requirements and latency that detector introduces. Without it,  the activations of 100 images from a single layer can take up to 22GB of memory. This aggregation step reduces the size of these vectors down to 0.6MB.

\textbf{Setting a Baseline.} To set a baseline for detection, we use points from the clean training dataset $X_{\text{clean}}$. For each layer $k$, we compute the mean $\mu_k$ and standard deviation $\sigma_k$ of the aggregate feature vector $A_k(\cdot)$ over the clean dataset $X_{\text{clean}}$. For each new input $x_i$, we compute its z-score vector $Z_k(x_i)$, where $Z_k(x_i) = (A_k(x_i) - \mu_k) / \sigma_k$ for each component in $A_k$, relative to the clean baseline vectors $\mu_k$ and $\sigma_k$. We use the top-$t$ percentile of the z-scores in the vector $Z_k(x')$ to identify the features that have the largest deviation from the baseline. Once we have the average $z_{\text{avg}, i}$ of the top-$t$ percentile values for input $x_i$ on layer $k$, we compare this to a threshold $\theta_{k,r}$ to determine whether or not to flag the input for that layer, where $\theta_{k,r}$ is chosen to have a desired False Positive (FP) rate $r$ over the trusted clean dataset. A flag is set to $\text{flag}(x_i) = 1$ to signal that input $x_i$ deviates from the baseline on layer $k$.

\textbf{Detector Calibration.} To calibrate the detector we need to choose: aggregation function $A$, layer observed $k$, and top percentile $t$. Given candidate layers $k \in K$ for observation, we perform the SGD-Layer attack for each $k$. For each of the generated SGD-Layer UAPs, we compute the AUC for both aggregation functions $M_k$, $S_k$ for each $k \in K$. The best layer  $k^\ast$ and aggregation method $A^\ast$ are chosen to maximize the average AUC for these attacks. The detector then observes the single layer that has the best aggregate performance. Grid search is used to find the best top-percentile value $t$.

\subsection{Measuring Detector Performance}
\label{sec:metrics}
The initial goal of our detector is to find inputs that have been perturbed by UAPs and to detect adversarial perturbations even if they have not been successful as attacks. This allows the defender to detect attempts to attack the model. Therefore we consider as true positives inputs that contain a UAP even if did not change the model's original prediction. We later introduce additional metrics to measure the effectiveness of the detector in practice.

To evaluate HyperNeuron, we take a subset of the training set as the trusted dataset $X_{\text{clean}}$ for computing the baseline vectors $\mu_k$, $\sigma_k$. Tests are then conducted on a separate validation set. We apply the UAP to samples from the validation set to measure the True Positive (TP) rate of the detector. The False Positive (FP) rate is based on the performance of the detector on the corresponding clean set of inputs. Details on the datasets used for each experiment are described in their corresponding subsections.

\textbf{Detectability-Universality Trade-Off.} We measure the detector's performance on layer $k$ on a test dataset using its AUC-ROC, which we shorthand with AUC. We then compare the detectability-universality trade-off for various UAPs based on their UER and AUC. Ideally, a good defense forces the attacker to trade-off UAP detectability and universality, making it more challenging to evade both the model and the detector on many inputs with a single UAP. We capture this relationship by comparing UAPs with their AUC and UER.

To measure the performance of HyperNeuron in practice, we assume that inputs flagged by the detector ($\text{flag}(x) = 1$) are discarded. Non-discarded inputs ($\text{flag}(x) = 0$) are classified by the model. Thus, we define two metrics: the Attack Success Rate (ASR) is the proportion of non-flagged perturbed inputs that have been misclassified, and the Clean Performance (CP) is the proportion of non-flagged clean inputs that have been correctly classified. The former penalizes the detector when it fails to discard misclassified perturbed inputs, while the latter penalizes the detector when it discards correctly classified clean inputs.

%% file: sections/4-imagenet.tex
\section{Large-scale Images}
In this section, we evaluate HyperNeuron's detection capabilities against adversarial mask and adversarial patch UAP attacks. This evaluation is done on the ImageNet \cite{russakovsky2015imagenet} dataset, a standard benchmark for computer vision tasks with 1,000 class labels.

\begin{figure}[t]
\centering
\includegraphics[width = 0.7 \columnwidth]{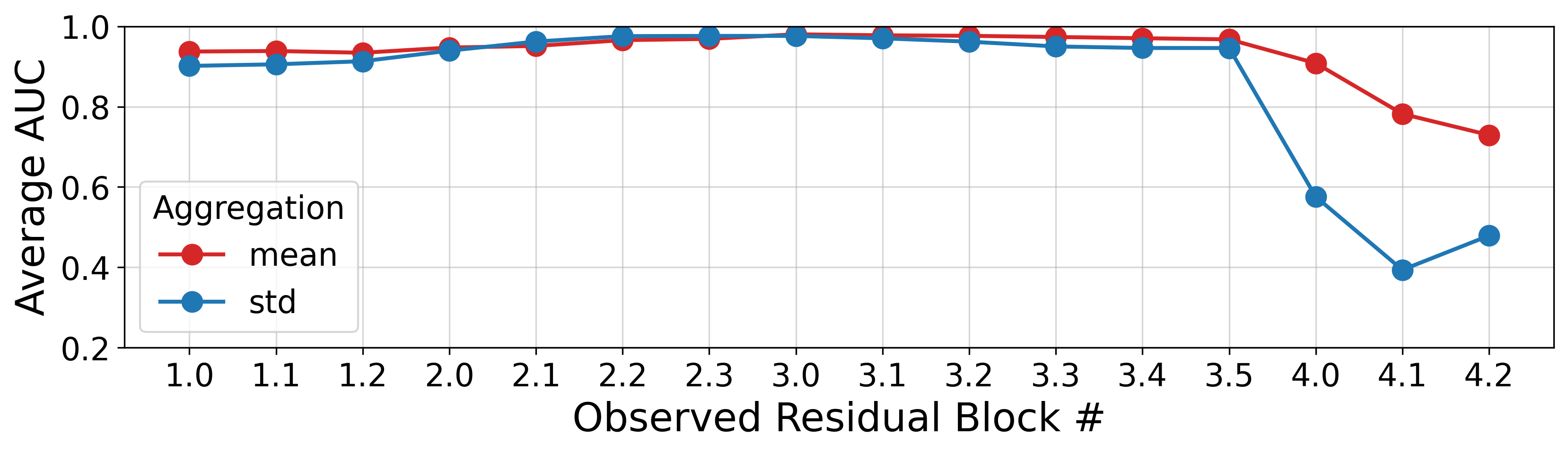}
\caption{Mean AUC for HyperNeuron with each combination of observed layer and aggregation function against SGD-Layer attacks with $\varepsilon = 10$ on ResNet50.}
\label{fig:in_calibration}
\end{figure}
\vspace{-2mm}

\begin{figure}[t]
\centering
\includegraphics[width = 0.77 \columnwidth]{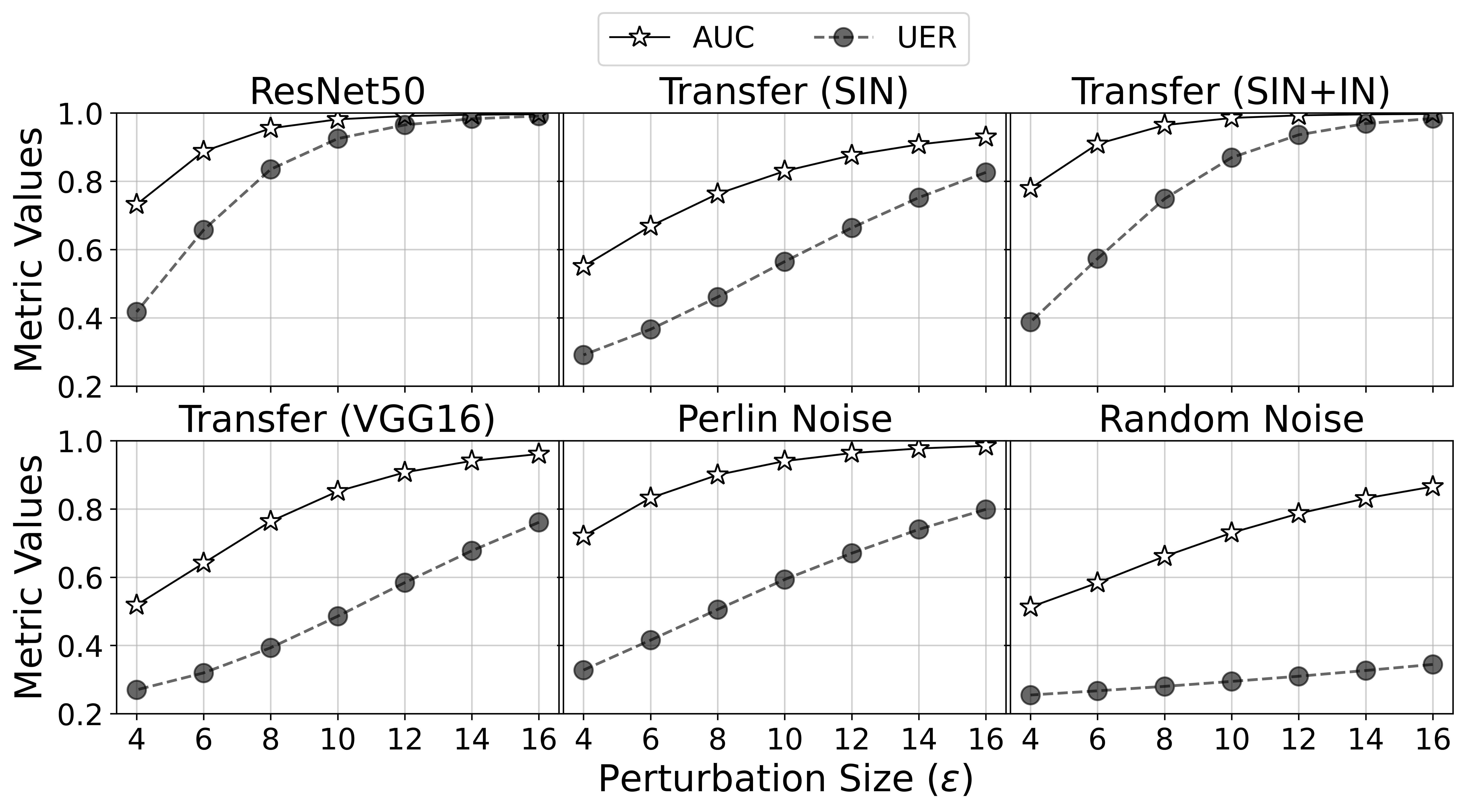}
\caption{Evaluation metrics of HyperNeuron - 3.0M on ResNet50 against untargeted direct and indirect attacks for different $\varepsilon$.}
\label{fig:in_uer}
\end{figure}

\begin{table*}[t]
\footnotesize
\caption{Metrics for targeted UAPs at $\varepsilon = 10$ against ResNet50 on ImageNet.}
\label{table:in_tgt}
\centering
\begin{tabular}{lccccc}
	Class ID & 698 & 109 & 815 & 854 & 611\\
	\hline
	Class Label & palace & brain coral & spider web & theater curtain & jigsaw puzzle\\
	\hline
	TSR (\%) & 70.60 & 83.35 & 83.41 & 85.04 & 86.33\\
	\hline
	Best AUC& 0.970 & 0.980 & 0.945 & 0.991 & 0.940\\
	\hline\\
\end{tabular}
\end{table*}

\subsection{Experimental Setup \& Calibration}
We evaluate HyperNeuron on ResNet50 \cite{he2016deep} which takes ImageNet images with dimensions $224 \times 224 \times 3$. Our chosen model achieves 76.13\% accuracy on the clean validation set and we calibrate HyperNeuron on the layer outputs of its 16 residual blocks numbered 1.0 to 4.2. Clean baseline vectors $\mu_k$, $\sigma_k$ for each layer are based on the activation values from 50,000 random training set images, with 50 examples for each of the 1,000 class labels. To determine the sensitivity of the detector to this baseline, we have repeated the evaluations for 10 different sets of training images and found no significant variance in detection performance.

\textbf{Attacks.} UAP evaluations are done on the 50,000 image validation set. For adversarial masks, we test on six different universal attacks, taking at least one representative example from each of the UAP types described earlier in Section~II: SGD (Untargeted, Transfer, Targeted), SGD-Layer, Perlin Noise. Then, we test our defense against adversarial patch UAPs \cite{brown2017adversarial}.

We first evaluate against direct attacks, as this is the most effective type and is often used in practice. It generates UAPs by directly optimizing over the model's loss, which can be adjusted for untargeted or targeted UAPs. For the transfer attacks, we generate untargeted SGD-UAPs for a VGG16 model \cite{simonyan2014very} trained on ImageNet, a ResNet50 model trained on Stylized-ImageNet (SIN) \cite{geirhos2019imagenet, co2019universal}, and a ResNet50 model trained on both Stylized-ImageNet and ImageNet (SIN+IN). Stylized-ImageNet is a variation of ImageNet using similar images, but designed to be differentiated based on high-level shape features rather than low-level textural features \cite{geirhos2019imagenet}.

For indirect attacks, we use Perlin Noise as the UAPs have greater universality, and we use Bayesian Optimization to find optimal attack parameters. We calibrate HyperNeuron on SGD-Layer as we can expect our detector to perform well against this attack.

\textbf{Calibration.} From the results shown in Fig.~\ref{fig:in_calibration}, we select the mean $M_k(\cdot)$ as our aggregation function and $k = 3.0$ as the layer to observe as this combination has the highest average AUC across the SGD-Layer attacks. We refer to this HyperNeuron configuration of layer 3.0 with mean aggregation as 3.0M.

\subsection{Adversarial Masks}
For adversarial mask UAPs, we see in Fig.~\ref{fig:in_uer} that UAPs with high universality (UER) are correlated with high detectability (AUC). And for smaller perturbation values, a lower AUC corresponds to a lower UER. The detection rates remain high for all attacks.

Among the different types of UAPs, direct attacks exhibit high detectability as they were generated on the same model. More surprisingly, transfer attacks also have high detectability, perhaps as a result of either the shared model architecture or training datasets. Indirect attacks suffer from low universality and require much larger perturbations to be effective. UAPs are naturally detected for larger perturbations as they drift from the baseline distribution. UAPs at lower perturbation values but with high UER are also detected often like the white-box SGD-UAP at $\varepsilon = 8$ with AUC = 0.95 and UER = 83.51\%. UAPs that are not as frequently detected tend to have much lower universality. These findings confirm a detectability-universality trade-off.

We sample targeted UAPs for 5 random class labels and report their performance and detection rates in Table~\ref{table:in_tgt}. Targeted UAPs with high TSR are all detected with AUC above 0.94. This demonstrates the efficacy of HyperNeuron against targeted UAPs.

\subsection{Adversarial Patches}
We evaluate against the ResNet50 model trained on ImageNet. In contrast to previous experiments, circular adversarial patches are constrained by a special case of $\ell_0$-norm and are also optimized to be robust to transformations. We assume that an adversarial patch covers less than a quarter of the original image.

To generate targeted adversarial patches we use a variation of the EOT objective as shown in Brown et al. \cite{brown2017adversarial}. We optimize the patches over 2,500 random images from the training set. This is sufficient to generate effective targeted adversarial patches that achieve at least 80\% TSR on the 50,000 image validation set when covering 16\% of the image area. We select three class labels to evaluate our defense against: \emph{jigsaw puzzle} (\#611), \emph{toaster} (\#859), and \emph{chocolate sauce} (\# 960). Toaster was chosen as it was the primary example used by Brown et al. \cite{brown2017adversarial}, jigsaw puzzle and chocolate sauce were chosen as they had the highest and lowest TSR amongst the targeted adversarial mask UAPs respectively.

\begin{figure}[t]
\centering
\includegraphics[width = 0.7 \columnwidth]{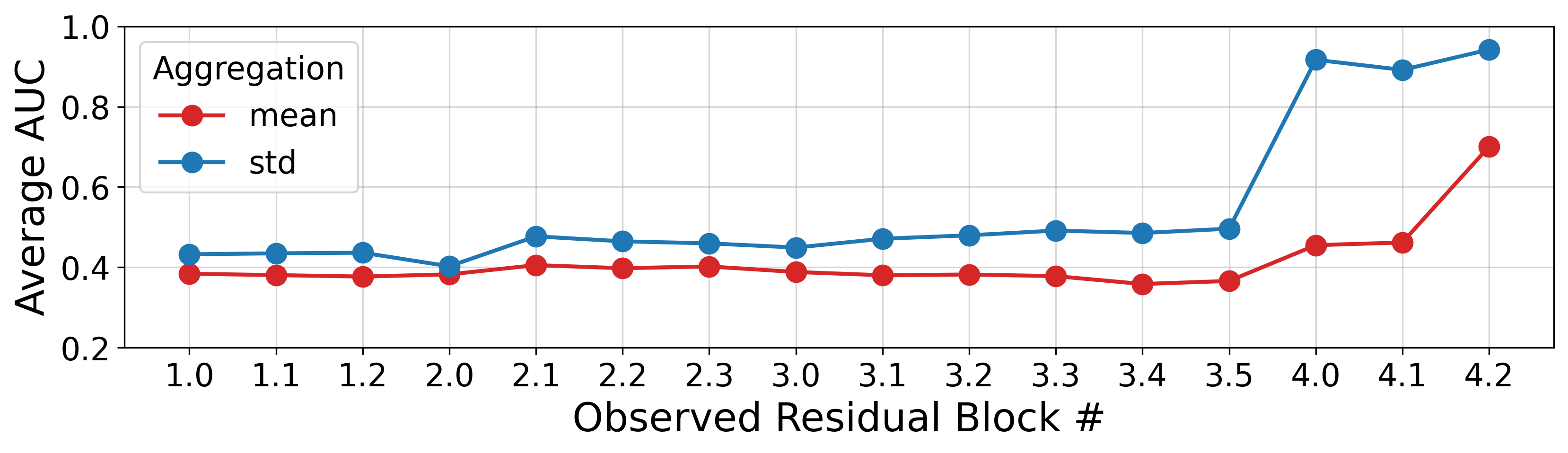}
\caption{Mean AUC for HyperNeuron with each combination of observed layer and aggregation function against Adversarial Patch attacks with 16\% patch size on ResNet50.}
\label{fig:in_calib-patch}
\end{figure}

\begin{figure}[t]
\centering
\includegraphics[width = 0.75 \columnwidth]{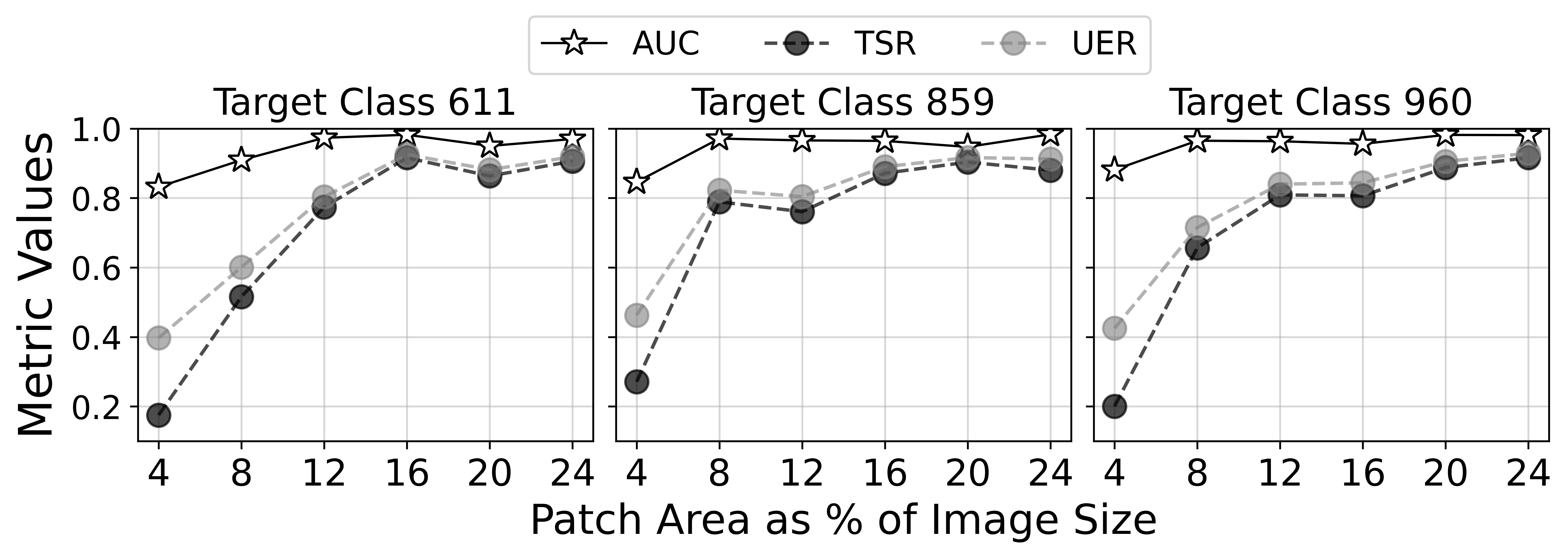}
\caption{Top row shows metrics for targeted adversarial patches against ResNet50 on ImageNet. There is significant overlap between the Best AUC and Block 4.2 AUC. Bottom row shows the AUC per layer.}
\label{fig:patch}
\end{figure}

\textbf{HyperNeuron Results.} Based on Fig.~\ref{fig:in_calib-patch}, we select the standard deviation $S_k(\cdot)$ as our aggregation function and $k = 4.2$ as the layer to observe as this combination has the highest average AUC. We refer to this HyperNeuron configuration as 4.2S. Compared to adversarial masks, patches are able to have larger pixel values but in more restricted regions of the image. Under these constraints, adversarial likely exploit higher-level features that are more robust to transformations and can be better recognized across images. It is these higher-level features that correspond to the final layers of the model. 4.2S is able to detect adversarial patches introduced with very high AUC (\textgreater 0.9) for all adversarial patches that cover 8\% or more of the image as seen in Fig.~\ref{fig:patch}. Even when a patch is not the most effective at 4\% of the image size, the patch is still reliably detected at 0.8 AUC or more. To conclude, our results show that both adversarial mask and patch UAPs with high UER are detected by HyperNeuron with high AUC. We next evaluate the performance of HyperNeuron with more practical applications in mind.

%% file: sections/5-practice.tex
\begin{table}[t]
\footnotesize
\caption{Metrics (in \%) summarizing average ASR and CP for HyperNeuron configurations against adversarial mask and patch UAPs on ResNet50.}
\label{table:cp}
\centering
\begin{tabular}{lcccc}
	& No Defense & 3.0M \, & 4.2S \, & 3.0M+4.2S\\
	\hline
    Avg. ASR: Mask & 80.27 & 14.43 \, & 78.89 \, & \textbf{13.52}\\
    Avg. ASR: Patch & 88.65 & 86.63 \, & 11.37 \, & \textbf{11.06}\\
	\hline
	Clean Performance & 76.13 & 73.55 \, & 74.17 \, & 72.91\\
	\hline
\end{tabular}
\end{table}

\begin{table}[t]
\footnotesize
\caption{Average time for inference and detection in milliseconds (ms) on a batch size of 64 images on ResNet50.}
\centering
\label{table:latency}
\begin{tabular}{lcccc}
	& No Defense & 3.0M & 4.2S & 3.0M+4.2S \\
	\hline
	Time per batch & 75.53 & 110.02 & 108.27 & 130.82\\
	Time per image & 1.18 & 1.72 & 1.69 & 2.04\\
	\hline
	Latency per image & - & 0.54 & 0.51 & 0.86\\
	\hline
\end{tabular}
\end{table}

\section{Detection in Practice}
We have shown that HyperNeuron is able to reliably detect adversarial mask and patch UAPs. Our results also show that UAPs with high universality often have high detectability with HyperNeuron. We consider now how to apply HyperNeuron in practice by: first, combining the detection for both mask and patch UAPs; and second, by measuring the latency our detector introduces.

\textbf{Combining Mask and Patch Detection.} We combine the detection of both mask and patch UAPs by flagging an input when it is flagged by either 3.0M or 4.2S. We refer to this configuration as 3.0M+4.2S. To measure the ASR and CP, we set the detector thresholds to have 5\% false positive rate on the clean baseline for each of 3.0M and 4.2S. Our results in Table~\ref{table:cp} show that this naive rule is sufficient to create a detector that reliably flags both types of UAPs.

The combined detection 3.0M+4.2S simultaneously reduces the average ASR on both mask and patch UAPs by 85\% compared to the ASR on no defense, while retaining more than 95\% of the original CP. When averaging over both mask and patch attacks, 3.0M+4.2S achieves much better performance only at a marginal cost of CP. Additionally, the layer that flags the input indicates to the defender whether the input is more likely to have a mask or a patch UAP.

\textbf{Runtime Analysis.} We measure the latency of HyperNeuron when defending the ResNet50 model from both adversarial mask and adversarial patch attacks in Table~\ref{table:latency}. The detection component of HyperNeuron is run on the CPU, and the model is run on a GeForce RTX 2080 Ti GPU.

Our results show that the runtime with HyperNeuron remains on the same order of magnitude as inference without the defense. This is in contrast to other defenses like SentiNet that introduce orders of magnitude more latency per batch at 2 seconds (2,000ms) \cite{chou2020sentinet} on similar hardware. The per-image latency that HyperNeuron introduces is a fraction of a thousandth of a second demonstrates that it is extremely efficient and suitable for real-time applications such as image classification or object detection in videos of up to 60 frames per second. The added 55 millisecond latency from HyperNeuron would then be negligible.

%% file: sections/6-comparison.tex
\section{Comparison with Existing Defenses}
Existing defenses for adversarial mask UAPs denoise the input or retrain the model to correct its output on perturbed inputs \cite{Akhtar_2018_CVPR, mummadi2019defending, shafahi2018universal, borkar2020defending}. These approaches are good for correcting model predictions on tested universal attacks without having to intervene during model inference. However, no information is gained as it is unclear if the errors made by the model are caused by maliciously altered inputs or by natural model error. HyperNeuron takes a different approach by performing detection instead. Detection is more advantageous as it alerts the user of active attacks, and helps identify input sources with compromised integrity.

\begin{table}[t]
\footnotesize
\caption{Comparison of HyperNeuron with non-detection defenses over mask and patch UAPs on ResNet models for ImageNet. Lower ASR and higher CP indicate better model performance. All metrics are in \%.}
\label{table:compare}
\centering
\begin{tabular}{lccccc}
	&& No Defense \, & SFR\cite{borkar2020defending} \, & AT\cite{wong2020fast} \, & M+S (Ours)\\
	\hline
    \multicolumn{2}{l}{Avg. ASR: Mask} & 80.27 & 29.77 & 48.72 & \textbf{13.52}\\
    \multicolumn{2}{l}{Avg. ASR: Patch} & 88.65 & 38.41 & 78.88 & \textbf{11.06}\\
    	\hline
	\multicolumn{2}{l}{CP} & 76.13 & 75.47 & 53.83 & 72.91\\
	\hline
\end{tabular}
\end{table}

\subsection{Comparison \& Results}
To compare HyperNeuron with non-detection defenses, we use Attack Success Rate (ASR) and Clean Performance (CP) metrics. For non-detection defenses, ASR is equivalent to UER and CP is the accuracy on the clean dataset. These metrics make sense as they measure the relevant performance of the model on the perturbed and clean datasets for both detection and non-detection defenses.

HyperNeuron is the first defense to demonstrate simultaneous efficacy against both adversarial mask and adversarial patch UAP attacks. However, we compare against defenses designed for each of the threats. For adversarial masks, we compare against Selective Feature Regeneration (SFR) \cite{borkar2020defending} and Adversarial Training (AT) as SFR outperforms other UAP defenses for recovering model accuracy on adversarial mask UAPs \cite{borkar2020defending}. AT is chosen as we have yet to see how adversarial training with attacks like Fast Gradient Sign Method (FGSM) \cite{goodfellow2015explaining} improves model robustness against UAPs. We use the fast AT algorithm \cite{wong2020fast} which trains against FGSM with $\ell_{\infty}$-norm = 4 on a ResNet50 model.

\textbf{Adversarial Mask.} Table~\ref{table:compare} shows that HyperNeuron (3.0M+4.2S) achieves much lower average ASR for both adversarial mask and patch attacks when compared to SFR and AT. HyperNeuron decreases the ASR closer to zero as it can reject perturbed inputs, whereas the other defenses are restricted by their error rate on the clean dataset. Existing non-detection defenses like SFR and AT attempt to restore the model's clean performance on the perturbed images. However, this means that the ASR of UAP attacks on these defenses would be at least their clean error on non-perturbed images. Thus, if the model's performance on the clean dataset is far from perfect (e.g. ResNet50 has 23.87\% clean error on ImageNet) then the ASR will be lower bounded by this value. This can be seen from our results in Table~\ref{table:compare}, where the ASR for any of the UAP attacks on SFR and AT always remain above their respective CE.

SFR is able to decrease the ASR of SGD-UAP much closer to its clean error. However Perlin Noise, which is a non-gradient based attack, achieves a better ASR than SGD-UAP. This shows that SFR does not generalize as well to unseen UAPs when compared to HyperNeuron. For mask UAPs against AT, we see in Table~\ref{table:compare} that their average ASR of 48.72\% is very close to the clean error of 46.17\%. This means that AT with FGSM does make the model insensitive to mask UAPs. However, as the CP is very low, the model still makes many errors on perturbed images. Additionally, AT does not improve the robustness against adversarial patch attacks. This is in line with recent results showing that AT defenses have difficultly achieving robustness against multiple types of perturbations simultaneously \cite{co2019procedural, kang2019transfer, tramer2019adversarial2}. Unlike SFR and AT, HyperNeuron does not attempt to recover the correct class label but can give information on the integrity of input sources. The automatic recovery of labels is an advantage if the model has near-perfect accuracy or if intervening with the deployed model is costly. Feedback on the integrity of inputs is useful if the defender actively monitors model inputs for malicious attacks or data drift. The defender has to consider that aspect  when evaluating and deploying a defense.

\textbf{Adversarial Patch.} We compare HyperNeuron with SentiNet, a UAP defense that detects adversarial patch UAPs \cite{chou2020sentinet}. Using model visualization and interpretability methods. Since their defense is a detector, we can directly compare the TP and FP rates on adversarial patch attacks. They report 98.5\% TP at 4.8\% FP for an adversarial patch that is roughly 25\% of the image area, which targets the \emph{toaster} (\#611) class on a VGG16 trained for ImageNet. In a similar attack on ResNet50, HyperNeuron achieves 94.9\% TP at 5.02\% FP for 3.0M+4.2S. SentiNet appears to achieve slightly better performance, but with a latency of 2 seconds on similar hardware. In contrast, HyperNeuron is primarily run on the CPU and introduces a drastically reduced latency of 0.055 seconds for a batch of 64 images. Thus, HyperNeuron remains orders of magnitude faster. SentiNet is not viable for real-time application. Whereas HyperNeuron at 0.055 seconds could be used for live detection.

\subsection{Conclusion}
Compared to existing defenses, HyperNeuron has the distinct advantage of being able to simultaneously defend against both adversarial mask and adversarial patch UAPs with significantly lower latency, while maintaining competitive or better ASR and CP. HyperNeuron has sufficiently low latency, to be easily applied to existing models without significantly increasing computation costs, and this makes it a practical choice for deflecting UAP attacks.

%% file: sections/7-adaptive.tex
\section{Defense-aware Adaptive Attack}
We consider the case where the attacker has the same capabilities as before, but now has knowledge of the defender's detection mechanism. The attacker's goal is to maximize ASR by generating a UAP that also evades HyperNeuron. ASR is an appropriate metric as it measures the proportion of unflagged misclassifications made on a perturbed set of inputs. UAP generation and evaluation is done on the 50,000 ImageNet validation set, and the detector thresholds are set to have 5\% FP rate on the clean baseline for 3.0M and 4.2S.

\begin{figure}[t]
\centering
\includegraphics[width = 0.8 \columnwidth]{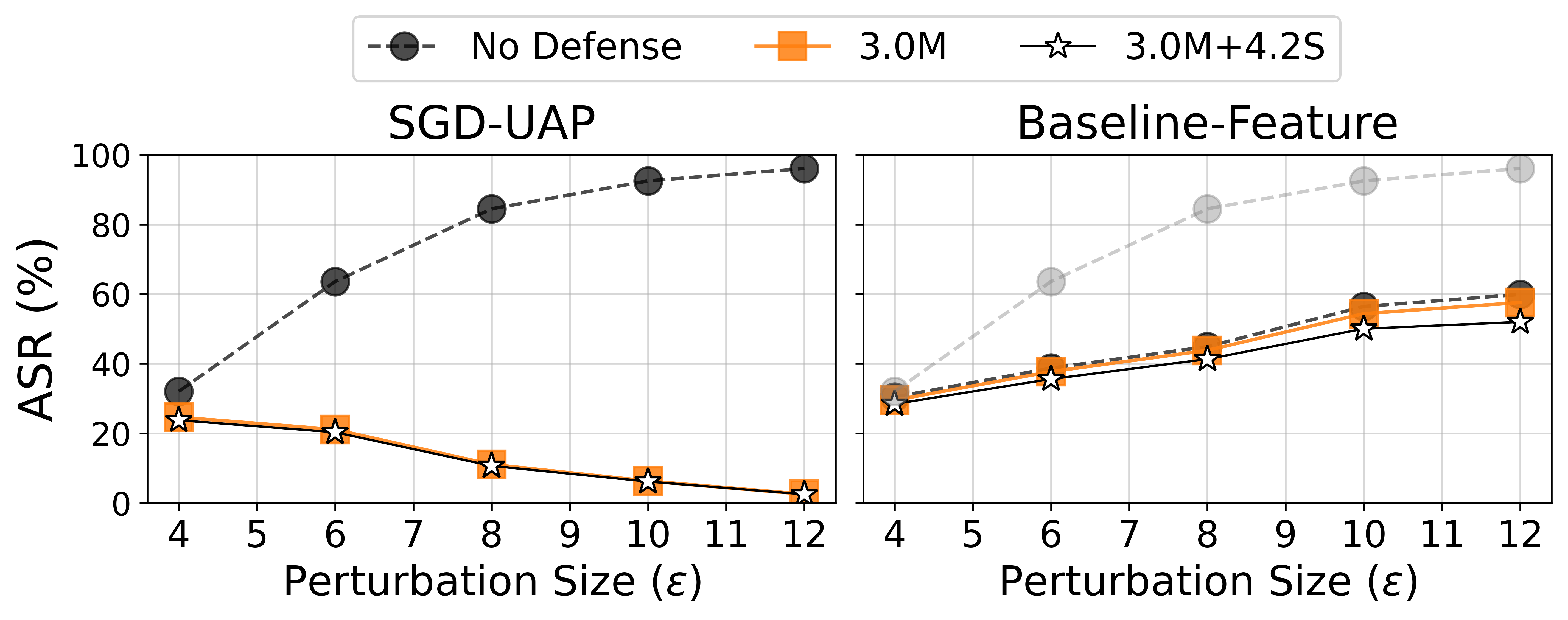}
\caption{Defense-aware attacks against HyperNeuron for ResNet50 on ImageNet validation. ASR of SGD-UAP on the undefended model is shown on the right subplot in gray for reference.}
\label{fig:in_adaptive}
\end{figure}

To generate our adaptive attack, we focus primarily on adversarial mask UAPs and assume the defender uses 3.0M for detection as it has shown to be the most capable for detecting mask UAPs. We apply SGD optimization to a new loss function which we call the Baseline-Feature attack:
\begin{equation}
\sum_i \left(\mathcal{L}(x_i + \delta) - \text{ReLU} \left(M_k (x_i + \delta) - \mu_k\right)\right)
\label{eq:adapt-uap}
\end{equation}
The first term is taken from SGD-UAP to maximize the model's loss, while the second term minimizes the resulting UAP's detectability. The ReLU penalizes the perturbation when its activation exceeds the value of the baseline vector, an assumption that HyperNeuron relies on.

Fig.~\ref{fig:in_adaptive} shows that the Baseline-Feature attack improves the overall ASR even when HyperNeuron is present. It is able to recover an average of 53\% of the original ASR when compared to SGD-UAP on an undefended model. However, the Baseline-Feature UAP generated has significantly less ASR on the undefended model (note ASR = UER when there is no detector). This shows that the detector still mitigates a large fraction of the UAP attack as it forces the attacker to sacrifice attack UER for a decrease in detectability, supporting the idea of the detectability-universality trade-off.

%% file: sections/8-adblocking.tex
\section{Application: Perceptual Ad-blocking}
Having seen the effectiveness of HyperNeuron against both adversarial mask and patch UAP attacks on ImageNet, we now test against perceptual ad-blocking as another example of a practical setting. We implement two UAP attacks against page-based perceptual ad-blockers and evaluate HyperNeuron on these attacks.

\begin{figure}[t]
\centering
\includegraphics[width = 0.77 \columnwidth]{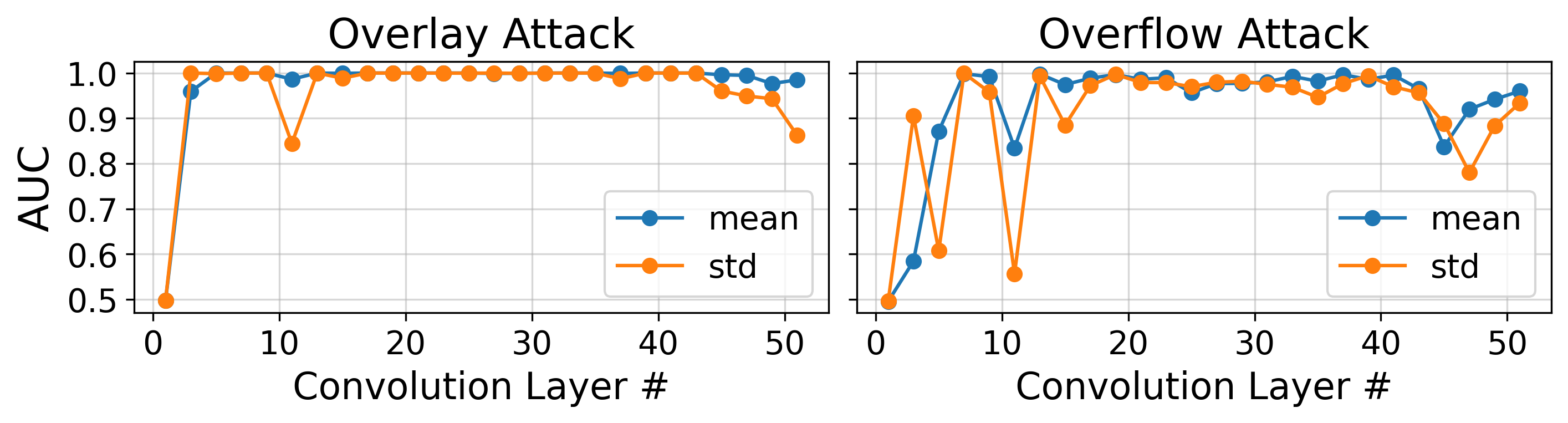}
\caption{AUC per layer of HyperNeuron for YOLOv3 against overlay and overflow attacks at $\varepsilon = 1$.}
\label{fig:ad_layer}
\end{figure}

\begin{figure}[t]
\centering
\includegraphics[width = 0.77 \columnwidth]{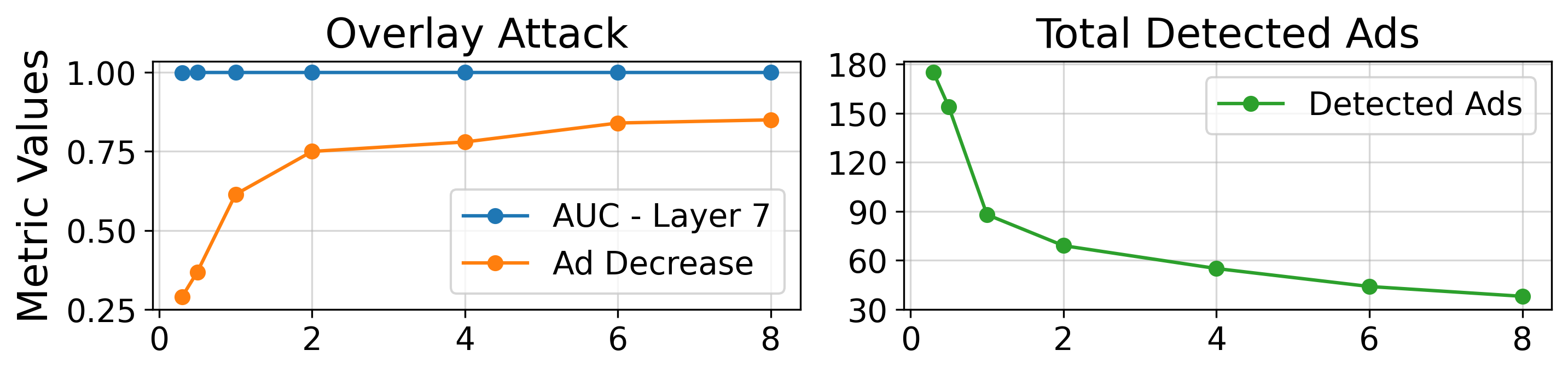}
\includegraphics[width = 0.77 \columnwidth]{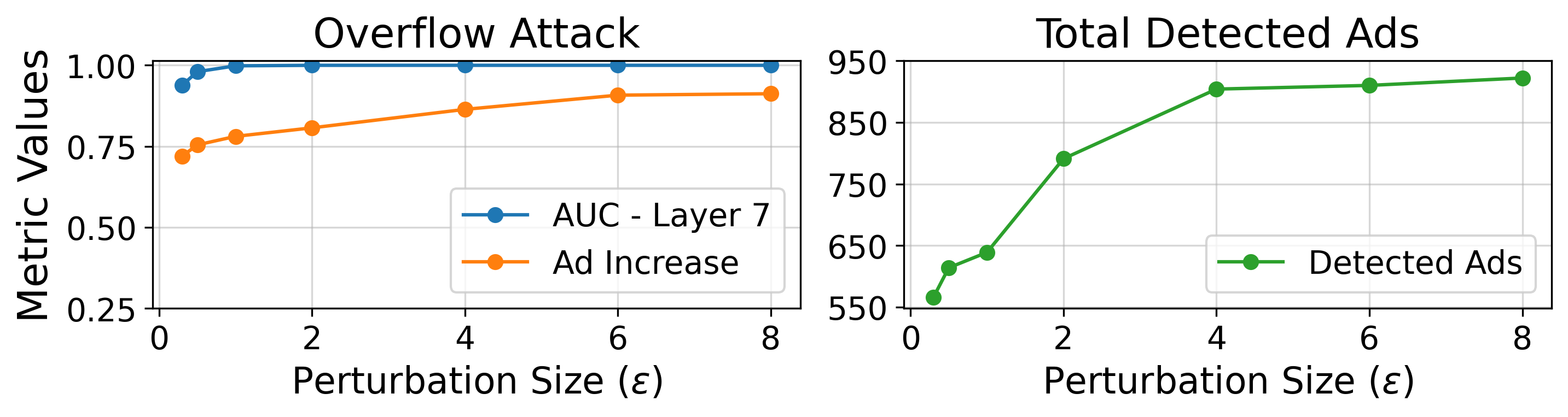}
\caption{Evaluation metrics for YOLOv3 against overlay and overflow attacks for different $\varepsilon$. The plots on the right show total number of detected ads.}
\label{fig:ad_uer}
\end{figure}

Perceptual ad-blocking detects online advertisements on a web page based on their visual content. However, Tramer et al. have shown that existing methods are vulnerable to UAPs \cite{tramer2019adversarial}. Here it is necessary for the perturbations to be \emph{universal} so that attacks are scalable and robust to arbitrary changes in the content of the web page. We take advantage of this property to detect the attacks.

\textbf{Model.} Page-based perceptual ad-blockers, such as \emph{Sentinel} \cite{adblock2020}, identify images on rendered web pages. This is an object detection model that outputs the bounding box coordinates of the detected ads and a confidence score. This is a binary task as there are only two classes (``ad'' and ``not ad''). The model used is based on YOLOv3 with a DarkNet53 feature extractor \cite{redmon2016you}, which has comparable performance to ResNet50 on ImageNet. We use the pre-trained object detector made available by Tramer et al. \cite{tramer2019adversarial}, which achieves an 80\% true positive rate for detecting ads on our test set of 225 web page screenshots.

\textbf{Attacks.} We focus on two types of attacks: the \emph{overlay} attack, where the UAP enables ads to evade detection, and the \emph{overflow} attack, where the UAP overflows the model with false ad predictions \cite{tramer2019adversarial}. Perturbations are generated by computing a small mask, then tiling the web page with the computed mask. SGD-UAP is used to optimize the mask for either hiding all ads or overflowing the model. These can be seen as targeted UAP attacks. We apply the two attacks on the test set of 225 web page screenshots for $\varepsilon = 0.3, 0.5, 1, 2, 4, 8$ under the $\ell_{\infty}$-norm perturbation constraints. For the overlay attack, we measure the total number of ads detected and ``ad decrease''--the percentage of pages where the UAP has decreased the number of ads detected. For the overflow attack, we measure the total number of ads detected and ``ad increase''--the percentage of pages where the UAP has increased the number of ads detected. These attacks are extremely effective even for very low perturbation values as seen in Fig.~\ref{fig:ad_uer}. The overlay attack halves the total number of ads detected at $\varepsilon = 1$ and the overflow attack doubles the total number of ads detected at $\varepsilon = 0.3$. These results demonstrate the fragility of the ad-blockers to visually imperceptible UAPs.

\textbf{Detection Results.} To detect the attacks, we monitor the convolution blocks of the DarkNet53. We see in Figs.~\ref{fig:ad_layer}, \ref{fig:ad_uer} the performance for each layer, and find that mean aggregation on layers 7, 13, 19, and 39 all achieve near perfect ($>$0.99) AUC in detecting both types of attacks, even for small perturbation values. HyperNeuron is able to detect all attacks with near-perfect AUC, as seen in Fig.~\ref{fig:ad_uer}. This shows that such detectors are highly vulnerable to perturbations but also highly detectable by HyperNeuron. Thus, in tasks where the universality of adversarial perturbations is important in generating realistic and successful attacks, we show that HyperNeuron is able to detect UAPs with near-perfect AUC for small perturbations.

%% file: sections/X-conclusion.tex
\vspace{-3mm}
\section{Conclusion}
UAPs can deceive machine learning models on many inputs and are used in practicals attacks. They appear to hyper-activate neurons to achieve high universality across inputs. Using this, we design HyperNeuron, a novel algorithm capable of successfully detecting UAPs including both adversarial masks and patches. We have tested HyperNeuron on the widely-used ImageNet benchmark and provide a comprehensive analysis on different types of UAPs attacks. We show that HyperNeuron is able to detect UAPs with high universality at high success rates. HyperNeuron runs with minimal latency of 0.86ms per image, greatly outperforms existing UAP defenses in terms of latency, while comparing favorably in terms of detection performance. It is also a general algorithm that can be applied to many tasks including image classification and object detection.

Finally, we highlight the application of UAPs to practical attacks where it is necessary for perturbations to be universal if they are to be robust to constant changes to the inputs. Beyond adversarial patches, we show that HyperNeuron achieves a high detection rate in a real use-case: perceptual ad-blocking. These results give us confidence that HyperNeuron puts defenses against UAPs on a more practical footing.